\documentclass[preprint,review,number,sort&compress]{elsarticle} 

\usepackage{subfig}
\usepackage{enumitem}
\usepackage{tabularx}
\usepackage{csquotes}
\usepackage{url}

\usepackage{xspace}
\newcommand{\totalsize}{40 million\xspace}
\newcommand{\filtersize}{2.6 million\xspace}
\newcommand{\archsize}{1 million\xspace}

\usepackage{amssymb}
\usepackage{xcolor}
\usepackage{etoolbox}
\definecolor{cb-green-sea}  {RGB}{  0, 146, 146}
\definecolor{cb-blue}       {RGB}{ 0, 109, 219}
\definecolor{cb-burgundy}   {RGB}{146,   0,   0}

\newcommand{\rXXX}{\textcolor{cb-burgundy}{\ensuremath{\blacklozenge}}\xspace} 
\newcommand{\rXX}{\textcolor{cb-burgundy}{\ensuremath{\blacktriangle}}\xspace}
\newcommand{\rX}{\textcolor{cb-burgundy}{\ensuremath{\blacktriangledown}}\xspace}
\newcommand{\rN}{\textcolor{cb-burgundy}{\ensuremath{\lozenge}}\xspace}

\newcommand{\bXXX}{\textcolor{cb-green-sea}{\ensuremath{\blacklozenge}}\xspace}
\newcommand{\bXX}{\textcolor{cb-green-sea}{\ensuremath{\blacktriangle}}\xspace}
\newcommand{\bX}{\textcolor{cb-green-sea}{\ensuremath{\blacktriangledown}}\xspace}
\newcommand{\bN}{\textcolor{cb-green-sea}{\ensuremath{\lozenge}}\xspace}

\newcommand{\sXXX}{\textcolor{cb-blue}{\ensuremath{\blacklozenge}}\xspace}
\newcommand{\sXX}{\textcolor{cb-blue}{\ensuremath{\blacktriangle}}\xspace}
\newcommand{\sX}{\textcolor{cb-blue}{\ensuremath{\blacktriangledown}}\xspace}
\newcommand{\sN}{\textcolor{cb-blue}{\ensuremath{\lozenge}}\xspace}

\begin{document}

\title{AI Art in Architecture}


\author[inst1]{Joern Ploennigs}
\author[inst1]{Markus Berger}

\affiliation[inst1]{organization={University of Rostock, AI for Sustainable Construction},
            addressline={Justus-v.-Liebig-Weg 6}, 
            city={Rostock},
            postcode={18059}, 
            country={Germany}}




\begin{frontmatter}
\begin{abstract}
Recent diffusion-based AI art platforms are able to create impressive images from simple text descriptions. This makes them powerful tools for concept design in any discipline that requires creativity in visual design tasks. This is also true for early stages of architectural design with multiple stages of ideation, sketching and modelling. In this paper, we investigate how applicable diffusion-based models already are to these tasks. We research the applicability of the platforms Midjourney, DALL-E 2 and StableDiffusion to a series of common use cases in architectural design to determine which are already solvable or might soon be. We also analyze how they are already being used by analyzing a data set of \totalsize Midjourney queries with NLP methods to extract common usage patterns. With this insights we derived a workflow to interior and exterior design that combines the strengths of the individual platforms.
\end{abstract}


\end{frontmatter}

\maketitle

\section{Introduction}\label{sec:intro}

Recent versions of AI art generation tools are reaching levels of output quality that allow them to support architects and designers in parts of their daily work. This gained them the attention of several architects across the globe to test out the applicability to their workflows reflected also in a special edition of the AEC magazine\footnote{\url{https://aecmag.com/technology/ai-special-edition-of-aec-magazine/}}.

Beyond this public discussion, we want to take a deeper look into the current capabilities of this technology and analyse qualitatively as well as quantitatively what benefits it offers to architects now and in the future. In this paper we therefore review the technology behind the three leading commercial AI art platforms and evaluate what use cases they currently support in architecture. We look into how users are already using these tools by analysing more than \totalsize public queries from one of these platforms with NLP workflows. Finally, we present a collection of case studies in which we apply the practical experience we collected in working with these systems.

The novel contributions of this paper are thus:
\begin{itemize}
    \item A comparison of the technology of three leading AI art platforms
    \item An analysis of how well current AI art tools can handle different use cases
    \item A mapping for what specific design tasks each platform supports
    \item An NLP analysis of how these platforms are used for architecture today
    \item A collection of practical workflows for specific architectural design tasks
\end{itemize}

\section{State of the Art in Generative Methods}\label{sec:soa}

The potential benefits of AI art platforms for creative work is hard to overstate. These AI art platforms are all using \textit{generative machine learning models}, specifically text-to-image generative models. Despite their specialization on generating images they are all based on the natural language model GPT-3 \cite{gpt-3}. GPT-3 is trained to generate text that completes a textual input query, more precisely it predicts the next possible combinations of words to an input text. The specific \textit{Image GPT-3} models used by AI art platforms are re-trained to instead predict the next cluster of pixels in an image for a given input text. 

The most recent generation of generative models combines natural language and so-called \textit{diffusion models}. The idea for diffusion models was first proposed in \cite{sohl2015deep}, in which (structured) image information is slowly destroyed through a \textit{forward diffusion process} that introduces noise into the image data and then generated anew through a \textit{reversed diffusion process}. This reverse process generates completely new image data, as the original information was fully destroyed by noise. This approach was constantly improved over the years with strong focus on optimizing the underlying neural network architectures \cite{ho2020denoising}.
There are several variants, like OpenAI's GLIDE model \cite{nichol2021glide}. It consist of an encoder that creates a text encoding based on the user prompt, a model implementing the diffusion based on this text encoding, as well as an upsampler that upscales and denoises the result. Current diffusion models often implement the process of text encoding and the association of those text encodings with image parts with the CLIP (Contrastive Language Image Pre-training) architecture presented in \cite{radford2021learning, ramesh2021zero} and used by DALL-E 1.

Perhaps the currently most advanced incarnation of the technology is DALL-E 2 (from here on out simply referred to as DALL-E), based on the unClip method developed in \cite{Ramesh22}. Here, an image encoder is introduced that encodes both text and images into a diffusion-based joint representation space (the \textit{prior}). The image generation is done by a similarly trained \textit{decoder}, which translates the prior's encoding back into an image. Another main platform in the field is the open source model StableDiffusion, which is also based on the CLIP text encoder. The third contender, Midjourney, does not published their models, but it is assumed that it is using a similar structure.

This kind of diffusion model architecture can fulfill a large array of image generation tasks. \textit{Image-to-image} models (img2img) take an image to create one or several new variants from it. This is usually a directed process, meaning that an intention is added by a user in the form of an additional text prompt. The words this prompt consists of act as a guideline for the image generation. If a certain part of the original image was deleted, the model can replace it with entirely new content based on the prompt. This approach is called \textit{inpainting} \cite{lugmayr2022repaint}. The same functionality can also be used for \textit{outpainting} (also called \textit{uncropping} \cite{saharia2022palette}) - extending an image outwards instead of filling in missing sections. If a user is requesting new additions or changes to the original image without manually deleting or masking out parts it is called \textit{image editing}. Image editing is not yet available in commercial AI art platforms, but there is recent work in single-image editing through text prompt, for example in Kawar et al. \cite{kawar2022imagic}. Is another diffusion step applied to add more details to the image in a higher resolution then it is \textit{up-scaled} (also called \textit{super-resolution} \cite{saharia2022image}). Is a completely new image generated from a user-written text prompt then a \textit{text-to-image} model (txt2img) instead of a img2img model is used. Often platforms offer multiple or all of these methods, with configurable weights between the individual images and words. 

Assessing the value of all these models and architectures is difficult. Attempts at quantitative evaluation are being made, for example in \cite{borji2022generated}. However, in such cases it is difficult to evaluate subjective metrics like style, i.\,e. whether an oil painting style looks better than a photorealistic one.

As for research into architecture-adjacent use cases, Seneviratne et al.\ \cite{seneviratne2022dalle} describe a systematic grammar for using DALL-E for the purpose of generating images in the context of urban planning. They open-sourced 12.000 images created by that grammar. They found that, though, many realistic images could be generated, the model has weaknesses in creating realistic real-world scenes with a high level of detail. 

There has also recently been progress in generating video \cite{ho2022video}, 3D models via point clouds \cite{luo2021diffusion,zhou20213d,zeng2022lion}, and even 3D animation data \cite{tevet2022human}, usually based on similar architectures to the image models or even directly involving an image model. While this is beyond the scope of this paper, especially the generation of 3D models could be a big opportunity for architecture. The 3D workflows could in fact look similar to the image-based workflows shown in this paper.

\section{Model Architectures and Interfaces}\label{sec:architectures}

Although the concept of generative art has been a research area for years, it only entered public perception with the advent of publicly available diffusion model platforms like DALL-E, Midjourney or StableDiffusion, which combine txt2img, img2img, inpainting and upscaling into an easy-to-use workflows.

These models are not only competing on the technical or quality level, but also in terms of user experience. Midjourney directly interacts with its community by sharing queries across public (or private) channels in the Discord messaging app. In contrast, DALL-E 2 is only accessible by individuals through a dedicated web interface with authoring and editing tools. StableDiffusion on the other hand is released as open source, which fosters a plethora of community-created tools that are used more than the official web-interface\footnote{\url{https://beta.dreamstudio.ai/}}.

\begin{figure}[ht]%
\centering
\includegraphics[width=0.9\textwidth]{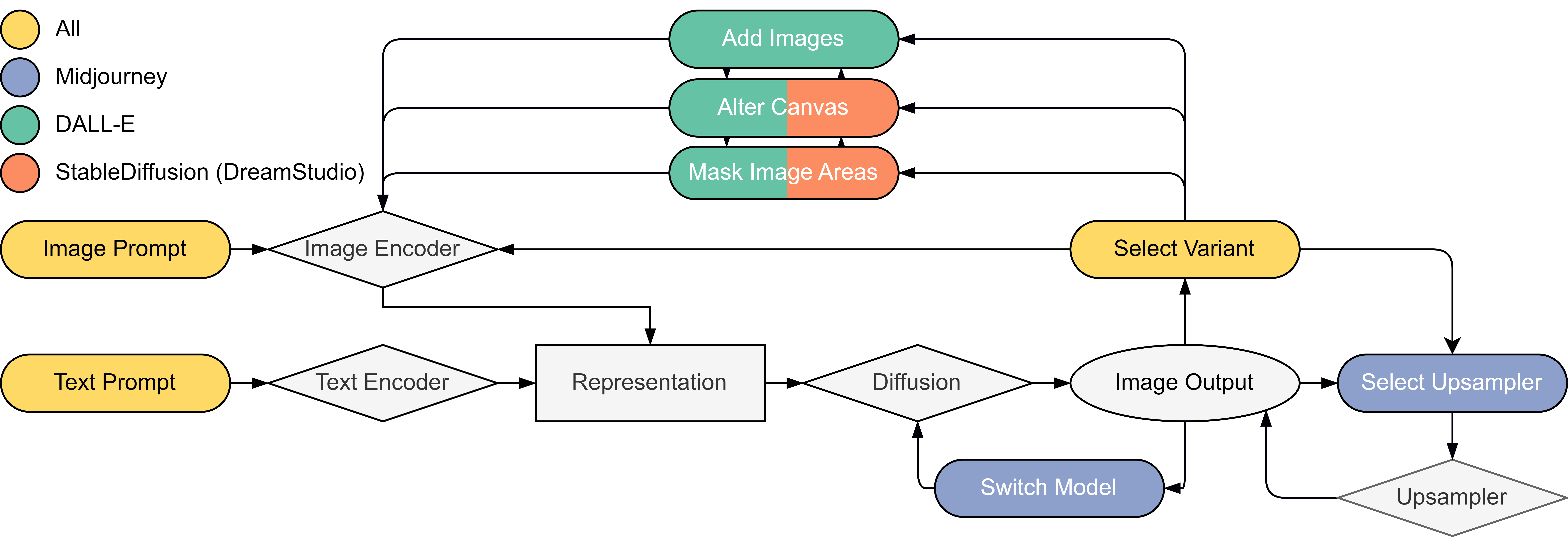}
\caption{Model architecture and image generation process in different models. Gray elements denote common model architecture steps, colored elements are user interaction points.}\label{fig:flow_all}
\end{figure}

The internal model architecture as well as the interface paradigm influences how these models can be utilized. In Fig.~\ref{fig:flow_all} we illustrate some of the similarities and differences between DALL-E 2, StableDiffusion, and Midjourney. The core blocks of foundational technologies in gray are similar across technologies as stated before. Differences lie in the workflows that they provide, which often results from their different interface approaches. 

It is apparent in Fig.~\ref{fig:flow_all} that while the core workflow in gray is similar, the ways to refine their results does vastly differ. Only Midjourney offers successive upscaling of resolution and does so with multiple different sizes. Thus, Midjourney's workflows center on generating and comparing image variants from different txt2img models and then upsampling the best one, with limited possibilities to remix the original text prompt.

In contrast, both DALL-E and StableDiffusion allow direct editing of uploaded images or txt2img results. They do not provide traditional image editing tools like drawing, filling, layering or stamping. Instead, all image editing has to be done by img2img-based operations like erasing sections (inpainting), extending the canvas (outpainting) or a combination of the two. All networks have ways to create images of different sizes and aspect ratios, either by specifying the size in the query or by altering it later through outpainting.

Notably, image generation only takes a few tens of seconds in all of the models, making it fast enough to use in creative sessions alone or with clients. All 3 models also allow importing external images in some capacity. Therefore, they can easily be combined into composite workflows, as shown in Section~\ref{ssec:exterior}.

One factor not included in Fig.~\ref{fig:flow_all} is the data that was used to train each model. There is little information on the training data used for DALL-E and Midjourney. However, StableDiffusion was trained on the LAION-5B dataset \cite{Rombach_2022_CVPR}, which is based on image and text data scraped from the web. Similar internet datasets are very likely the source for Midjourney and DALL-E. However, it is evident that either biased by the training data or the training process, these models have developed very different image styles. DALL-E and StableDiffusion have no trouble generating both drawn images as well as photorealistic outputs. Midjourney's tends towards a more impressionistic style. This can be altered through targeted keywords, but, even explicitly realistic results in Midjourney tend to be more abstract until the recently added \enquote{remaster} model variant is explicitly invoked on a previously upscaled image. With carefully chosen inputs most styles can be targeted, however, for ease of use it is often better to pick a platform that is already good at generating the desired style.

\section{Architectural Use Cases}\label{sec:usecases}

Given the discussed differences of the platforms in user interface and technology, they vary strongly in the architectural use cases that they support. 

To analyze this, we collected a series of use cases where architects normally create or edit images. A direct comparison between the platforms and the use cases is often complicated, because of their style variants. We identified that the more differentiating aspect in most cases is, whether or not a platform supports a specific technical feature that is required to realize the use case.

Therefore, we evaluated for each use cases what features they require and how qualitatively well this currently works across supported platforms. We also map which platform supports which feature fully or in limited capability. In addition to the image operations explained in the previous chapter, we also consider support for architectural semantics, i.\,e.\  structured knowledge that goes beyond common image training datasets. Table~\ref{tab:usecases_comparative} shows the results for the use cases that we discuss below:

\begin{itemize}
    \item \textit{Ideation:} Developing ideas by randomly generating images for inspirations. This is what txt2img models are made for and it works splendidly. Additional image prompts can add style and object references.
    \item \textit{Sketches:} Drawing architectural sketches with specific target style and items. This works well for common examples in the training data, but, less so with specific requests.
    \item \textit{Collages:} Combining and filling existing images with life by adding people and objects. This can be done through inpainting for individual items, but, not generic requests like: "Add many people".
    \item \textit{Image Combination:} Taking multiple existing image elements (for example multiple buildings), arranging them on a canvas and then creating a coherent composite image.
    \item \textit{Build Variants:} Taking an existing sketch or picture and generating versions in which certain elements are altered (like adding a garage). This works well trough inpainting.
    \item \textit{Style Variants:} Taking an existing image and transforming its style (e.g. a sketch to photorealistic art deco) without changing content. This works well with certain models.
    \item \textit{Construction Plans:} Creating detailed layout plans to establish spatial relations. This rarely works as the model do not understand the semantics of line styles, areas, etc.
    \item \textit{Exterior Design:} Finding style and feeling for a building and the surrounding area/landscape. This works well for common scenarios.
    \item \textit{Interior Design:} Finding a style or feeling for an interior space. This also works well in many scenarios.
    \item \textit{Creating Textures:} Creating tiled patterns to serve as surface materials for 2D or 3D models. This is currently a unique feature of Midjourney.
\end{itemize}

\begin{table}[ht]
\centering
\begin{small}
\setlength\tabcolsep{2pt}
\begin{tabular}{l|cc|cccc}
Model                    & txt2img      & img2img       & In-/Out-paint. & Editing      & Upscaling     & Semantics\\
\hline
DALL-E                   & \rXXX        & \rXXX         & \rXXX         & \rN           & \rN           & \rN  \\
Midjourney               & \rXXX        & \rXX          & \rN           & \rN           & \rXXX         & \rN  \\
Stable Diffusion         & \rXXX        & \rXXX         & \rXX          & \rN           & \rN           & \rN  \\
\hline
Use Case \\
\hline
Ideation                 & \bXXX/\sXXX  & \bX/\sXXX     & \bN/\sX       & \bN/\sN       & \bN/\sXX      & \bN/\sN\\
Sketches                 & \bXX/\sXX    & \bX/\sXX      & \bX/\sX       & \bN/\sN       & \bN/\sXX      & \bX/\sN \\
Collages                 & \bN/\sN      & \bX/\sN       & \bXX/\sXX     & \bXXX/\sN     & \bN/\sN       & \bN/\sN\\
Image Combination        & \bN/\sN      & \bXX/\sN      & \bXXX/\sXXX   & \bX/\sN       & \bX/\sN       & \bN/\sN \\
Build Variants           & \bN/\sN      & \bXX/\sX      & \bXXX/\sXX    & \bXXX\sN      & \bN/\sN       & \bXX/\sN\\
Style Variants           & \bN/\sN      & \bXXX/\sXX    & \bN/\sN       & \bN/\sN       & \bX/\sX       & \bN/\sN \\
Construction Plans       & \bXX/\sN     & \bN/\sN       & \bN/\sN       & \bXX\sN       & \bXX/\sN      & \bXXX/\sN \\
Exterior Design          & \bXX/\sXX    & \bX/\sX       & \bXX/\sXX     & \bXX/\sN      & \bX/\sX       & \bX/\sN\\
Interior Design          & \bXX/\sXX    & \bX/\sX       & \bXX/\sXX     & \bXX/\sN      & \bX/\sX       & \bX/\sN\\
Creating Textures        & \bXX/\sXX    & \bN/\sX       & \bN/\sN       & \bX/\sN       & \bXXX/\sXX    & \bN/\sN
\end{tabular}
\caption{Top part - Comparison of platforms with their supported features with: \rXXX full, \rXX limited, \rX bad, or \rN no support. Lower part - Mapping of architectural use cases to features with: \bXXX high, \bXX some, \bX low, or \bN no importance; versus (/) how well it works: \sXXX well, \sXX somewhat, \sX a little, \sN not at all.}
\label{tab:usecases_comparative}
\end{small}
\end{table}

Even though StableDiffusion seems to support fewer features than DALL-E, its main advantage can't be overstated: It is possible to run it locally and add additional training data to the model to introduce it to new architectural concepts. Together with the high quality of StableDiffusions outputs, this makes it the most potent of the three models to be specialized on architectural drawings.

\section{Analysis of Architectural Queries}\label{sec:inuse}

We also explored how people use these AI art platforms in practice. We analyzed about \totalsize queries of Midjourney users, as it is the only system for which many user queries are publicly visible. Midjourney uses the Discord messaging app as its main interface, which allowed us to monitor the public channels for queries that we consider to be of an architectural nature. 
We selected queries containing either the word \enquote{architect}, \enquote{interior} or \enquote{exterior} design or one of 38 architectural keywords like \enquote{building}, \enquote{facade}, or \enquote{construction}. We identified these keywords by selecting only those from architectural glossaries\footnote{\url{https://www.heritage.nf.ca/articles/society/architectural-terms.php}}$^,$\footnote{\url{https://en.wikipedia.org/wiki/Glossary_of_architecture}} that co-occurred in at least 10\,\% of all cases with \enquote{architect}, \enquote{interior} or \enquote{exterior} in the queries. We also added to the list of keywords the names of 941 famous architects from Wikipedia\footnote{\url{https://en.wikipedia.org/wiki/List_of_architects}} as we noted that several queries actually refer to the style of these architects. By applying these filters, we identified \filtersize queries (6.8\,\%) with potential architectural intent including \archsize queries (2.6\,\%) explicitly containing \enquote{architect}, \enquote{interior} or \enquote{exterior} design.

In the next steps we filter out stopwords and build a Word2Vec model  \cite{mikolov2013distributed} from these queries to get a model of the occurrence and co-occurrence of terms in these queries. For understanding the results, it is important to know that most Midjourney users do not formulate full sentences, but, rather a collection of terms that refer to the content, style, or render quality of the targeted image.

\begin{figure}[t]%
    \centering
    \subfloat[\centering Word Frequency]{{\includegraphics[width=0.33\textwidth]{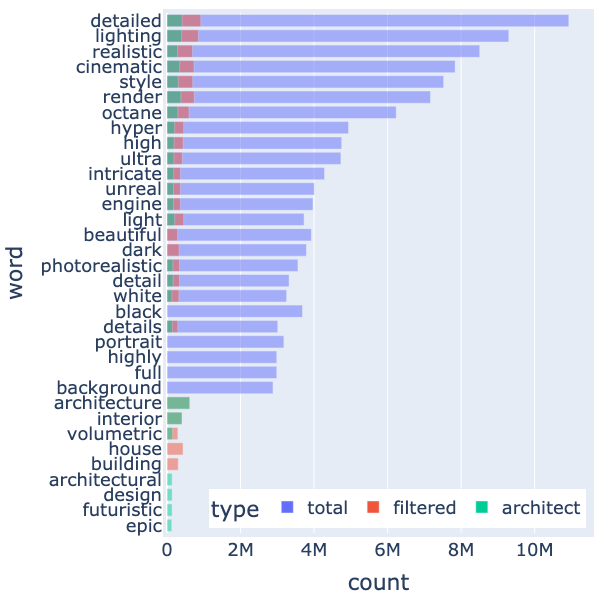} }}%
    \subfloat[\centering Keyword Frequency ]{{\includegraphics[width=0.33\textwidth]{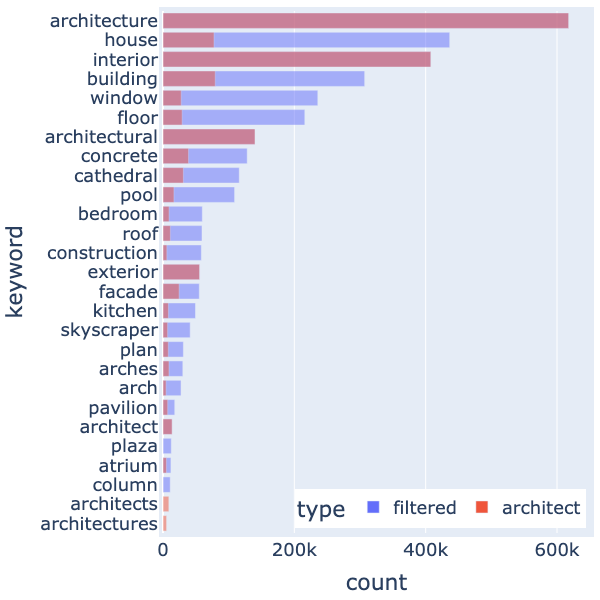} }}%
    \subfloat[\centering Famous Architect Frequency ]{{\includegraphics[width=0.33\textwidth]{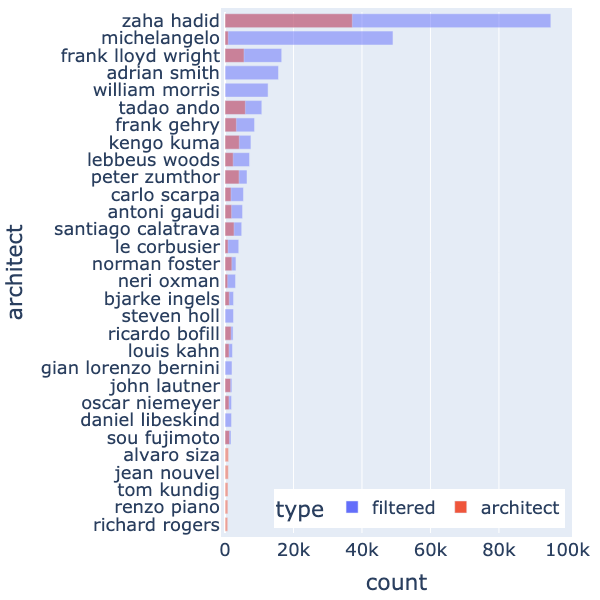} }}%
    
    \subfloat[\centering Keywords (left) and co-occurring terms (right) ]{{\includegraphics[width=0.33\textwidth]{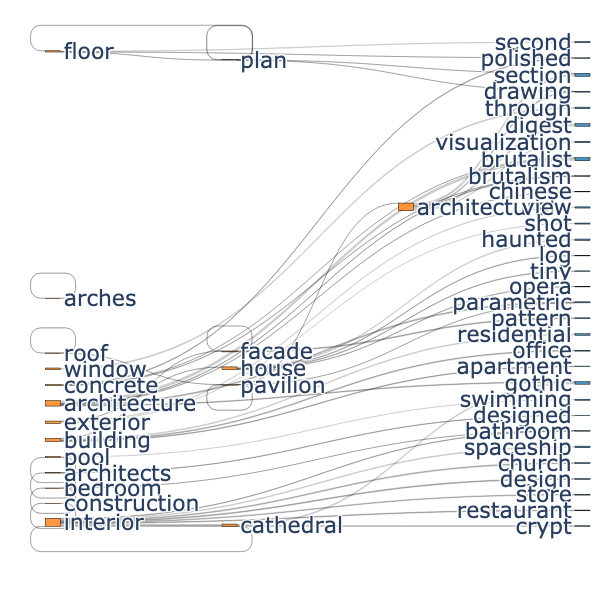} }}%
    \subfloat[\centering Query Length. x-axis labels show the \# of queries.]{{\includegraphics[width=0.33\textwidth]{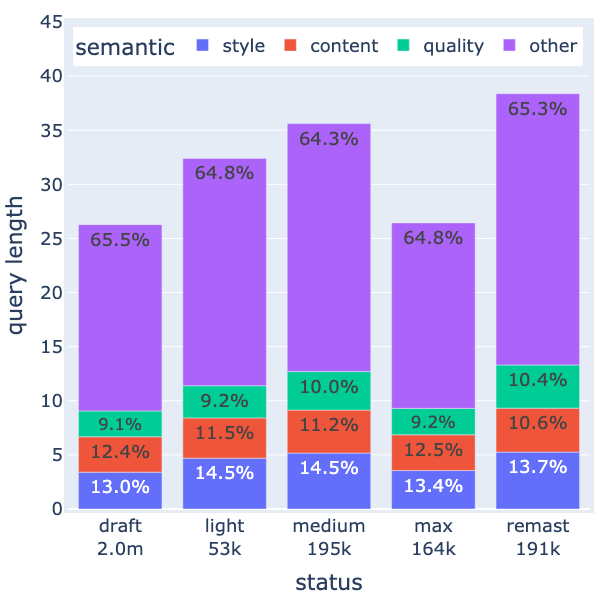} }}%
    \subfloat[\centering Query Workflows. x-axis labels show the \# of flows.]{{\includegraphics[width=0.33\textwidth]{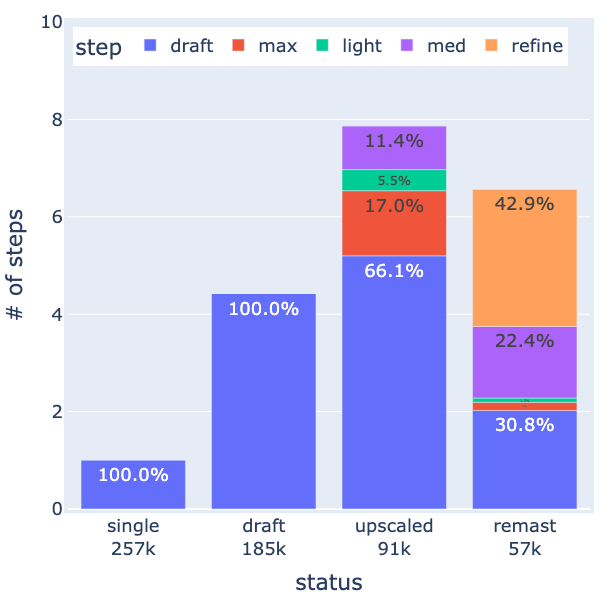} }}%
    \caption{Results from analysing Midjourney queries}%
    \label{fig:word2vec}%
\end{figure}

Fig.~\ref{fig:word2vec} visualizes the main results of our analysis. Fig.~\ref{fig:word2vec} (a) shows the  most frequent terms used in the filtered queries. The color blue represents the frequency across all \totalsize queries, red is the frequency within the filtered \filtersize queries and green within \archsize queries explicitly containing \enquote{architect}, \enquote{interior} or \enquote{exterior} design. The top 10 has a similar frequency across all three classes. Many of those refer to Midjourney style commands like \enquote{detailed}, \enquote{realistic}, \enquote{cinematic}, \enquote{render}. However some terms like \enquote{black}, \enquote{full} or \enquote{portrait} have high overall frequency, but, low frequency in architectural context. Other terms like \enquote{architecture}, \enquote{interior}, \enquote{house}, \enquote{building} do only occur exclusively within our filtered results as they are part of our keyword list.

Fig.~\ref{fig:word2vec} (b) lists these keywords that we use and their respective frequency. As we filter on these keywords, their total frequency is identical with the filtered one and we do not display it. It is of note that \enquote{architecture} and \enquote{interior} keywords are the most and third frequent keyword.

Fig.~\ref{fig:word2vec} (c) shows the frequency of famous architects that we extracted from 263.041 queries that referred to at least one of them. Zaha Hadid is by far the most frequently queried architect, given her well recognizable style and the likely popularity of her designs within the same social groups that are interested in experimenting with AI tools. 

Michelangelo is second, usually invoked for his art style instead of his architectural contributions, similar to William Morris. The architects Frank Lloyd Wright, Tadao Ando, Frank Gehry, Lebbeus Woods, Kengo Kuma, and Peter Zumthor, and Antoni Gaudi complete the top 10 and are often used in an explicitly architectural context.

Fig.~\ref{fig:word2vec} (d) shows the links between keywords and the most likely connected term. We analyzed this by predicting with the Word2Vec for each keyword on the left the most probably co-located word on the right, weighted by probability. Interesting combinations here are links between interior-appartment, floor-plan, architecture-parametric, arches-gothic, or pavilion-roof-flat. From this it is possible to build an of auto-complete function for architectural queries.

Fig.~\ref{fig:word2vec} (e) shows the mean length of queries depending on whether they got upscaled, remastered or left in draft mode. A draft mode image is of low image size, so users will normally upscale or remaster them if they like one of the variants. It is notable that for the upscale options (light, medium/beta) as well as the remastered version the mean query length increases above 33 terms per query in comparison to 26 terms for draft mode queries (interestingly this does not generalize to max upscale). We also manually classified the most frequent 150 terms into the categories: style, content, quality. It is notable that for the upscale and refined queries, the percentage of style terms increases significantly.

Fig.~\ref{fig:word2vec} (f) shows the mean number of iterations needed to develop a query. We classify a query as iteration if the same user is rerunning the same or extended query within 30 minutes. 9.84\,\% of all queries (single, 257k) are run only once. The maturity of queries is improved in multiple iterations. Queries that remain in draft mode require about 4.4 steps. Queries that are good enough to be upscaled require about 7.8 steps in total. They are upscaled after 5.1 draft steps into different variants (light, medium/beta, max). Queries that are remastered take about 6.5 iterations. They have only 2.0 draft mode queries, but 2.8 remastering steps, and 1.5 final upscale steps. 

This illustrates that users do not come-up with perfect queries from scratch, but normally develop them over multiple iterations by selecting the best variants or adding more terms (especially style terms). From these insights as well as practical experience we developed some recommended workflows that we will discuss in the next section.

\section{Case Studies}

In this section, we present some concrete workflows that the three AI art platforms enable architects to take. These workflows are based on the results of our analysis and the experience we have gained by running a large number of experiments on their interfaces. As we have seen, we rarely run only a single query and gain a perfect result - instead we iterate many times. If one does not enter into this process with a robust idea of appropriate steps, it is easy to query into dead ends and run out of compute credits quickly. 

\subsection{Interior Design - Comparing the Models}

First, we will take a look at how the models perform purely on their own. The example will be an interior-design scenario. We start with a simple query that does not use any of the special commands and a parameters of the platforms. On all three we try to generate a high-quality rendering of a room characterized as \enquote{cozy living room, wood paneling, television, large sofa, natural light, lived in, realistic, full view}.

\begin{figure}[t]%
    \centering
    \subfloat[\centering Midjourney original query ]{{\includegraphics[width=0.25\textwidth]{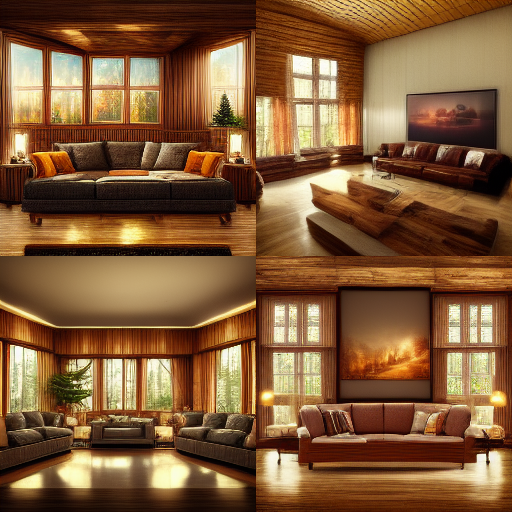} }}%
    \subfloat[\centering Variant selection ]{{\includegraphics[width=0.25\textwidth]{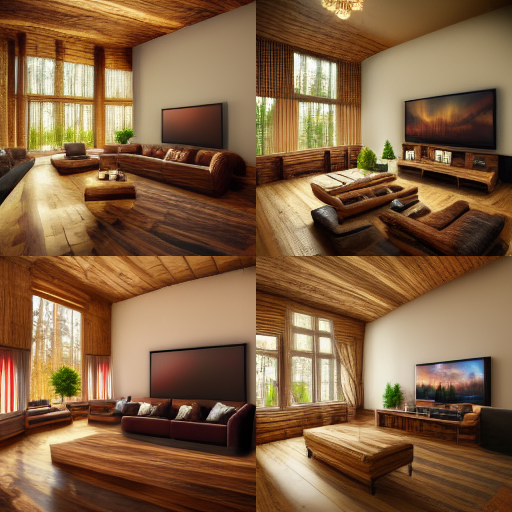} }}%
    \subfloat[\centering Upscale ]{{\includegraphics[width=0.25\textwidth]{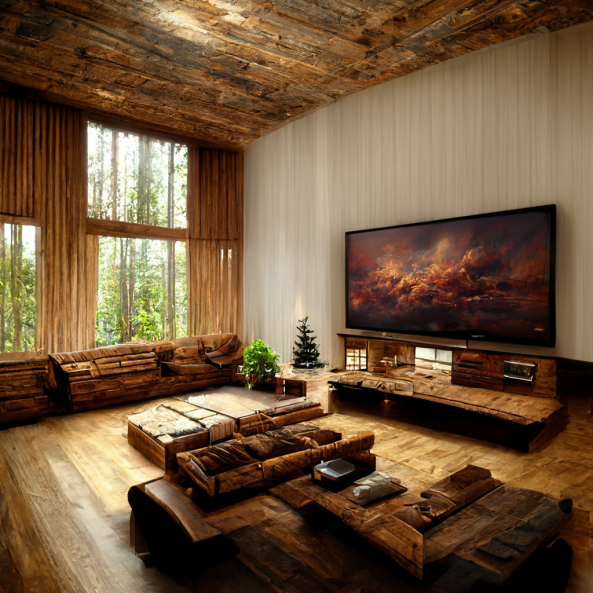} }}%
    \subfloat[\centering Remaster and maximum upscale]{{\includegraphics[width=0.25\textwidth]{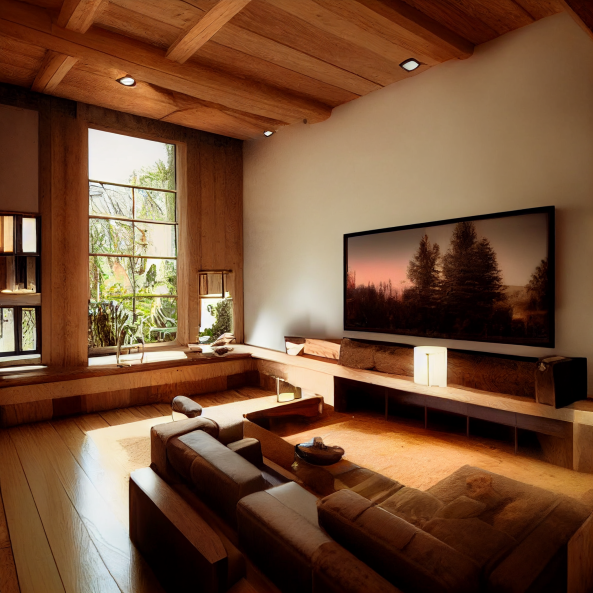} }}%
\vspace{-0.1cm}
    \subfloat[\centering DALL-E original query]{{\includegraphics[width=0.25\textwidth]{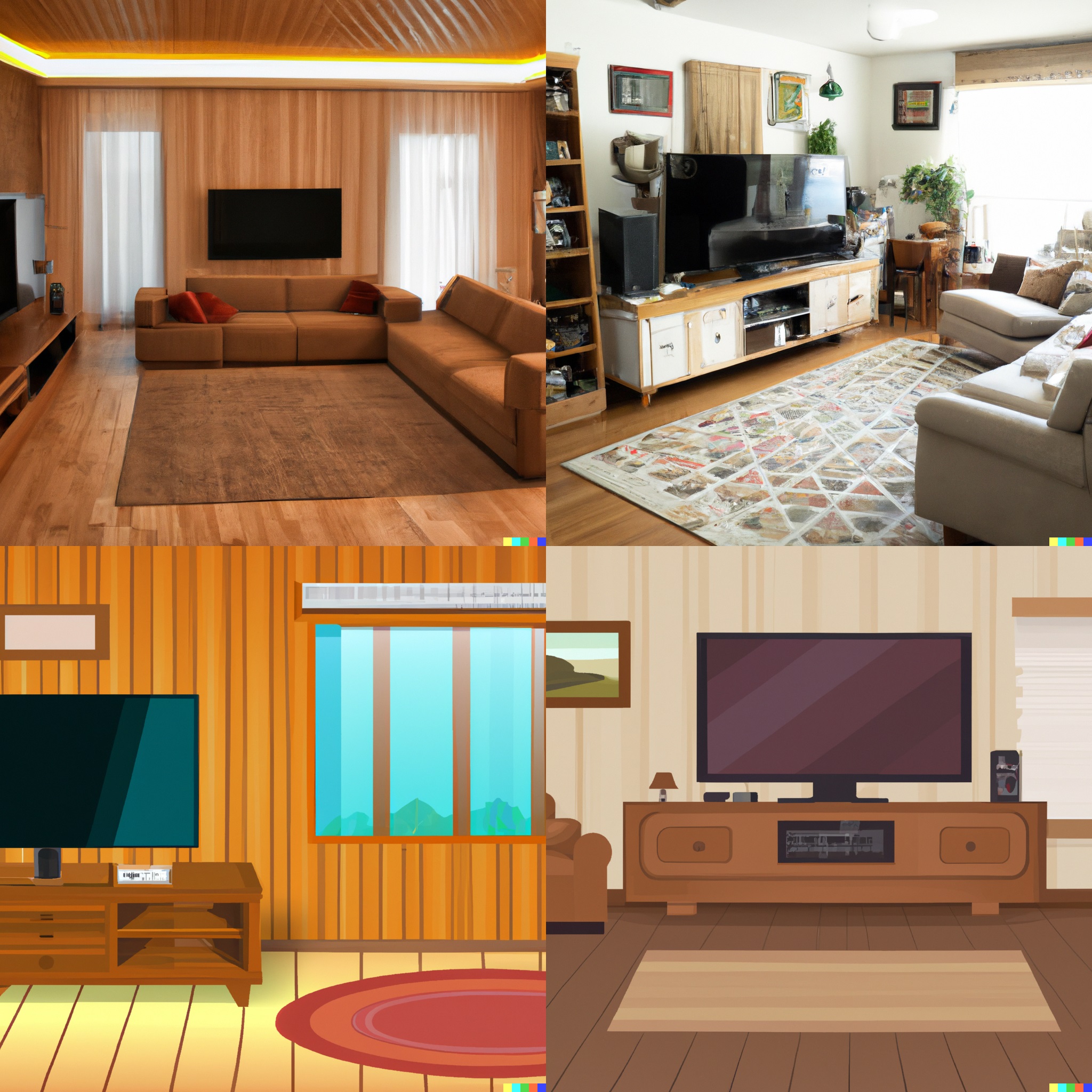} }}%
    \subfloat[\centering Variant selection]{{\includegraphics[width=0.25\textwidth]{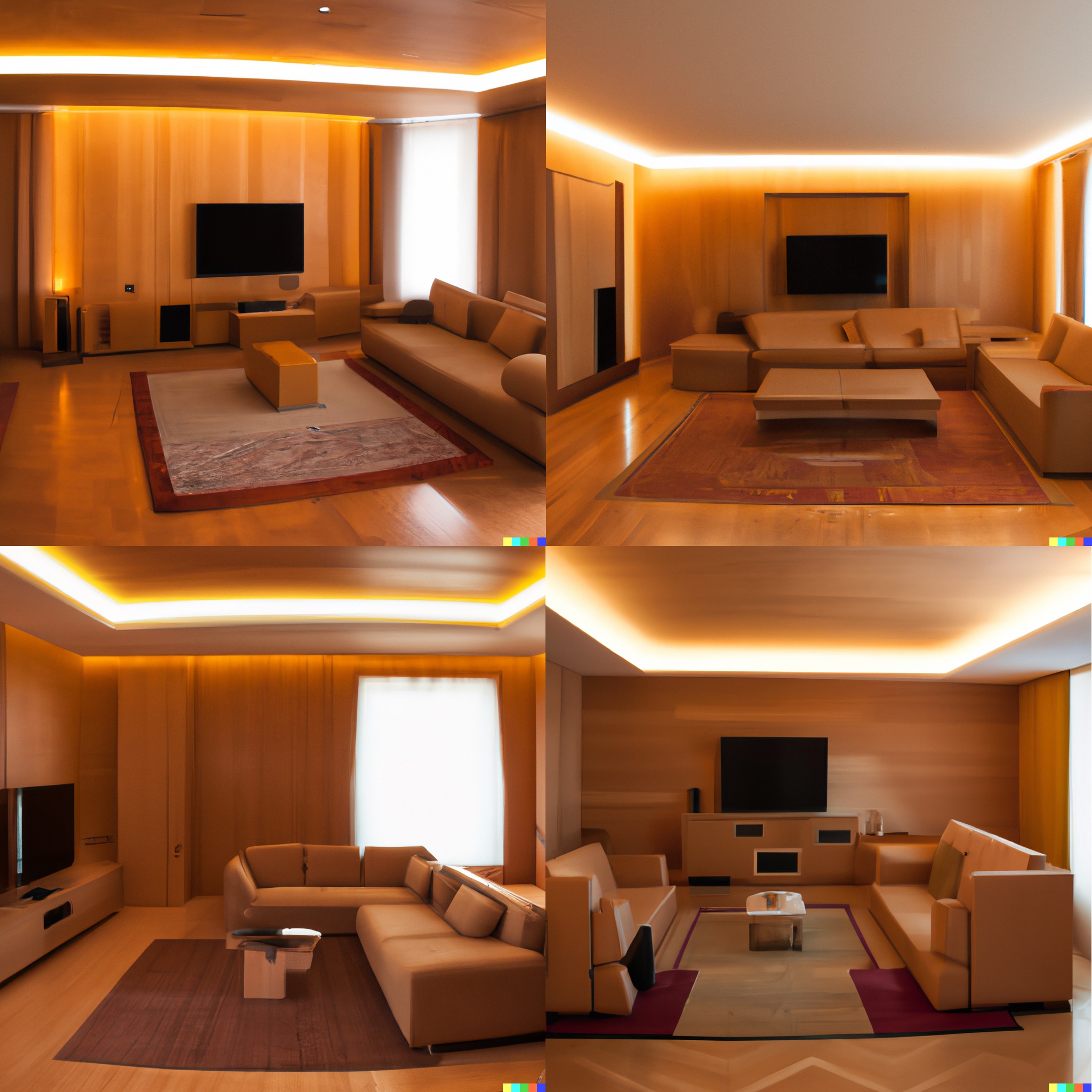} }}%
    \subfloat[\centering Erase tool to remove arti\-facts]{{\includegraphics[width=0.25\textwidth]{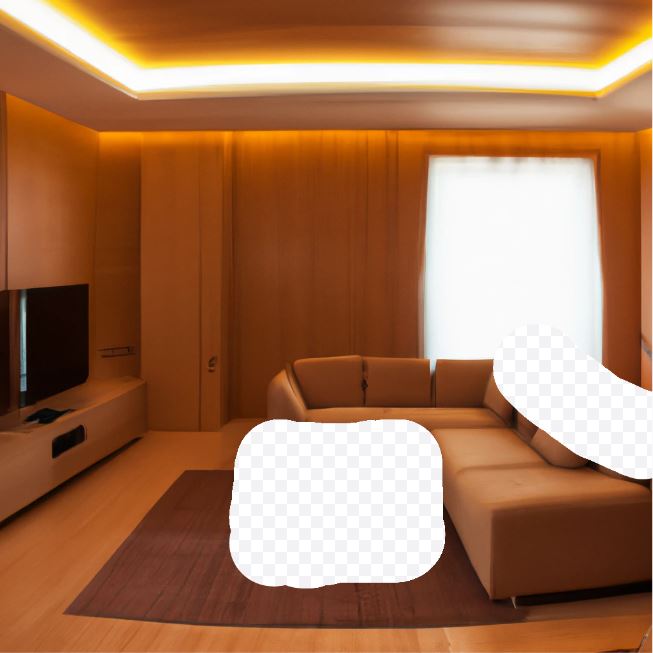} }}%
    \subfloat[\centering Final inpaint with added \enquote{table}]{{\includegraphics[width=0.25\textwidth]{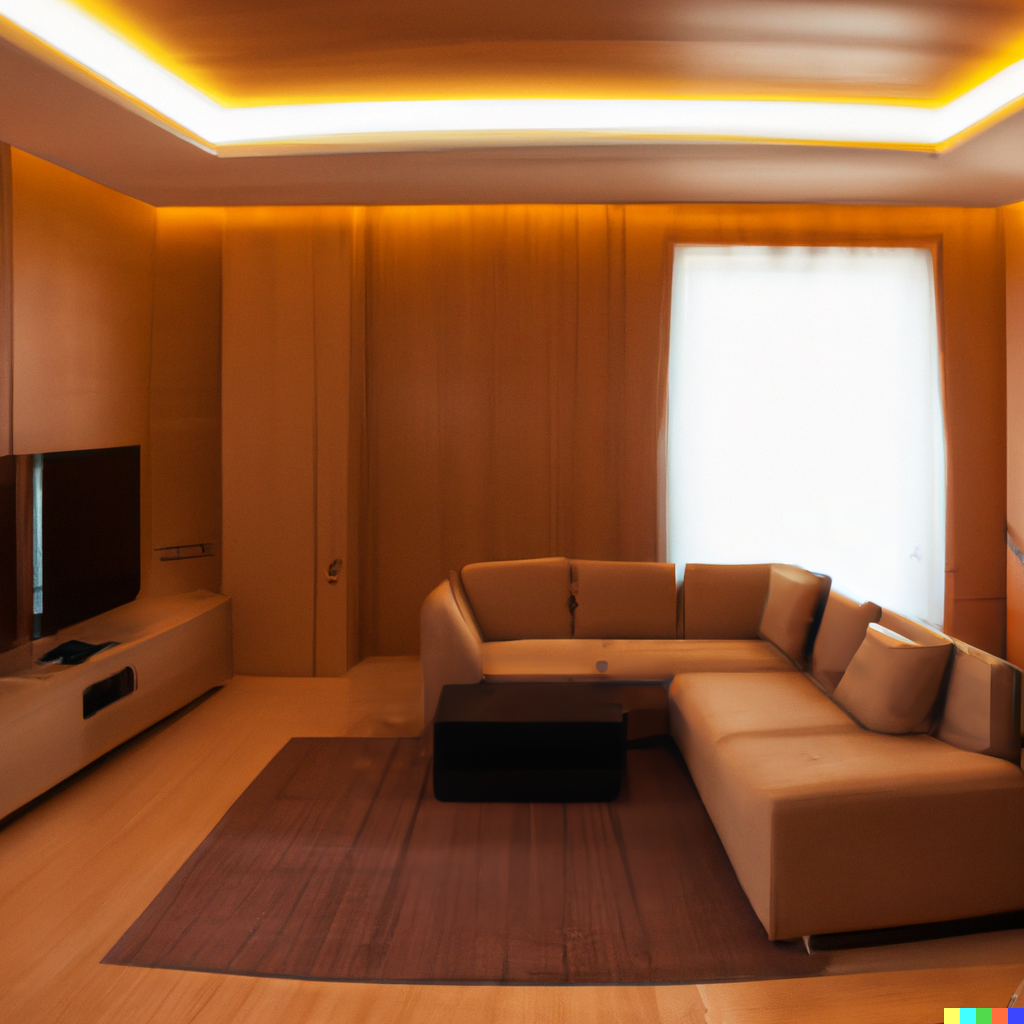} }}%
\vspace{-0.1cm}
    \subfloat[\centering Stable Diffusion original query]{{\includegraphics[width=0.25\textwidth]{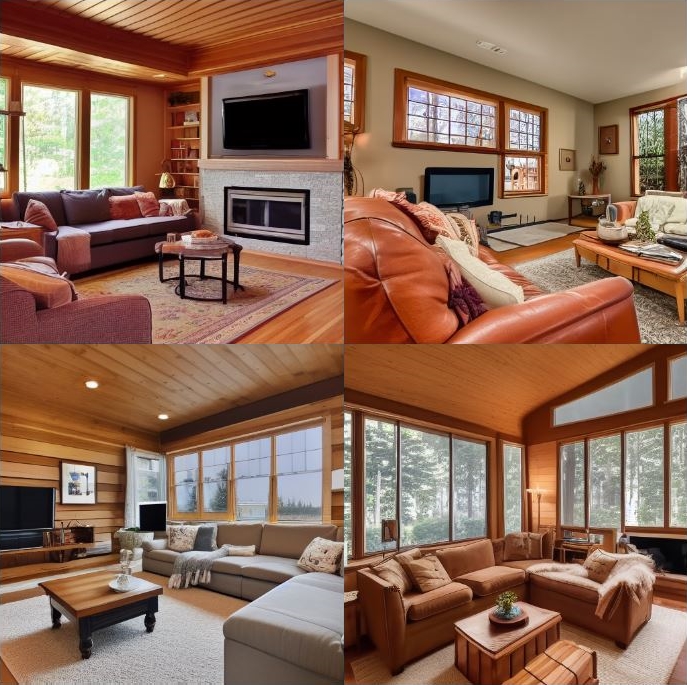} }}%
    \subfloat[\centering Resize selection and Erase artifacts]{{\includegraphics[width=0.25\textwidth]{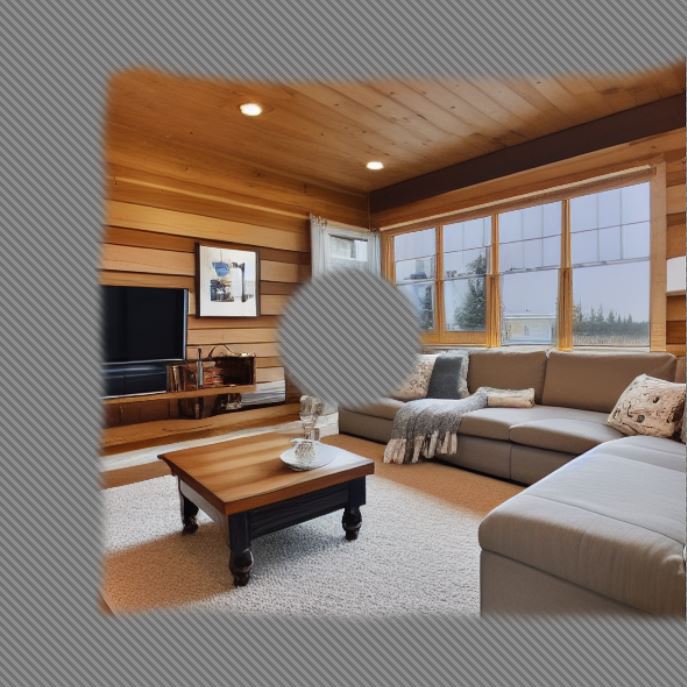} }}%
    \subfloat[\centering Variants of in-/outpainting]{{\includegraphics[width=0.25\textwidth]{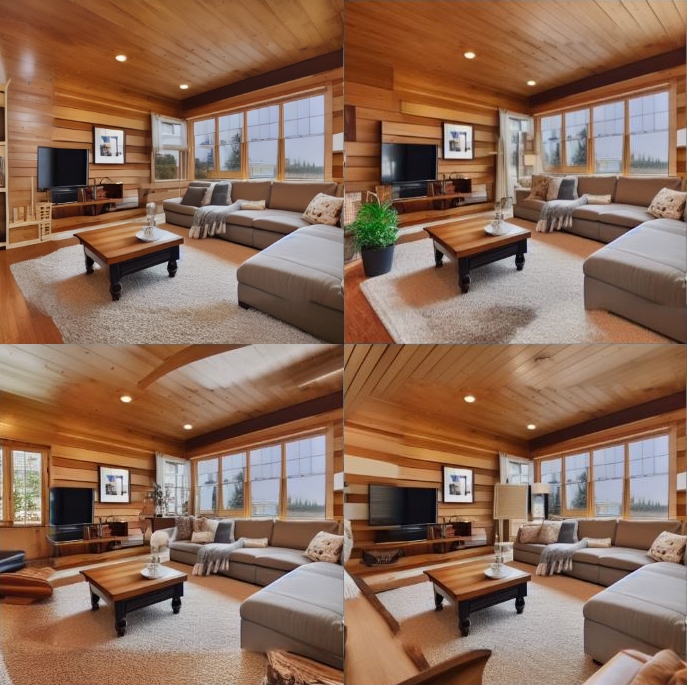} }}%
    \subfloat[\centering Final selection]{{\includegraphics[width=0.25\textwidth]{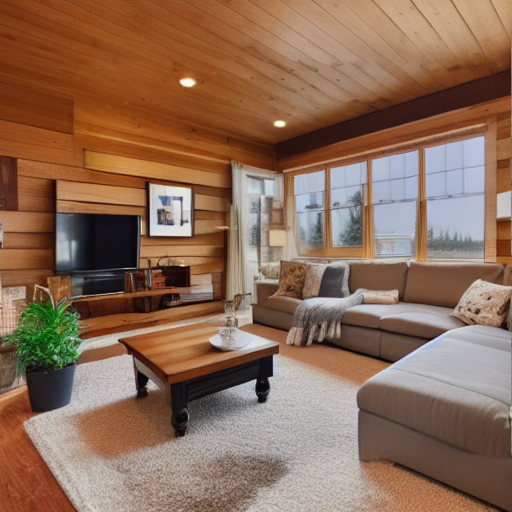} }}%
\vspace{-0.1cm}
    \caption{Minimal workflow for Midjourney (a)-(d), DALL-E 2 (e)-(h), and Stable Diffusion (i)-(l) for the given query.}%
    \label{fig:interior}%
\end{figure}

Midjourney starts out with several results that are stylized or of strange perspective, visible in Fig.~\ref{fig:interior} (a)-(b). The first upscale of the chosen interior design in (c) greatly improves material quality and overall detail, but the central sofa remains as an incoherent form in the center of the image. Anytime the normal image output does not attain a sufficient level of cohesion and realism, we can invoke the \enquote{remaster} step, shown upscaled in (d). However, even this last result contains smaller perspective errors, which are difficult to fix without intervention through manual image editing.

The DALL-E 2 first results in Fig.~\ref{fig:interior} (e) shows that it struggles to correctly response to the \enquote{realistic} term in the query. Once one of the two realistic variants is picked, the next variant generation step in (f) creates more useful results. To remove perspective or coherence errors we mask out certain areas of the image in (g) to generate inpaint variants. The final variant is shown in (h). Of note is that even the inpainted regions react correctly to the previously established lighting of the scene. This result is of good quality, but can't be upscaled any further within the web interface.

Stable Diffusion starts out in Fig.~\ref{fig:interior} (i) with much stronger results than its two contenders, generating images that incorporate the \enquote{realistic} and \enquote{lived in} aspect of the query very well. These rooms look like they are inhabited and not like artificial renderings. However all images are close-ups for a proper interior view. Which is why we not just use inpainting in (j) to fix errors, but, also add additional canvas space for outpainting. This results in some quite incoherent variants for the outpainted areas. After some additional iterations the variant in (l) was selected as the best.

Overall, StableDiffusion performs in this scenario best as all the beginning variants are quite realistic. It also shows no real deficiencies in the later steps. One only needs to watch out during the final outpainting stage, where it may be needed to manually smooth the transitions into the outpainted areas. Midjourney generated a final result of similar quality, however offered a weaker beginning selection for free design ideation.

\subsection{Exterior Design - Combining the Models}\label{ssec:exterior}

In order to play to the strengths of each model, one has to learn the terms and parameters they are trained to operate on and what style of image they are truly supposed to generate properly. One of the first hints that DALL-E shows new users is the information that the keyword \enquote{digital art} can significantly improve many prompts, which is deeply related to the data it was trained on.

In the case of Midjourney, refinement starts as a process of including and removing certain phrases within the prompt to get as close to a desired style as we can. These phrases can be very convoluted. It often helps to include the kinds of modelling software that would create the desired kind of image (like \enquote{octane render} or \enquote{cinema3D}) or even quality signifiers like \enquote{top 10 on artstation} into the prompt. A much more directed way to influence queries is the use of weighting words, image references and parameters, which add additional parameterization to the prompt system.

\begin{figure}[ht]%
\centering
\includegraphics[width=0.9\textwidth]{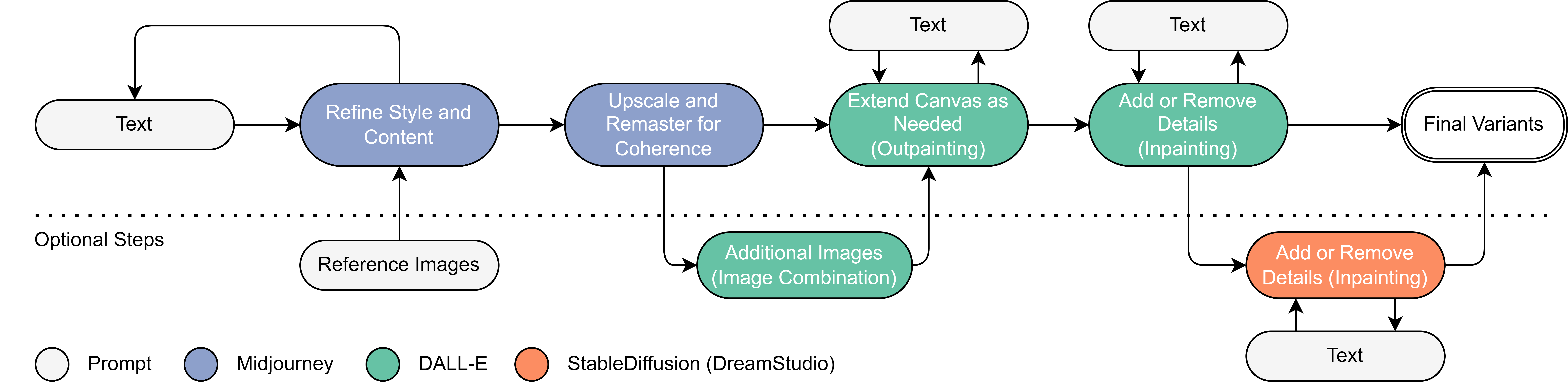}
\caption{The proposed combined workflow over Midjourney, DALL-E and (optionally) StableDiffusion.}\label{fig:combination_workflow}
\end{figure}

\begin{figure}[t]%
    \centering
    \subfloat[\centering Midjourney original query]{{\includegraphics[width=0.33\textwidth]{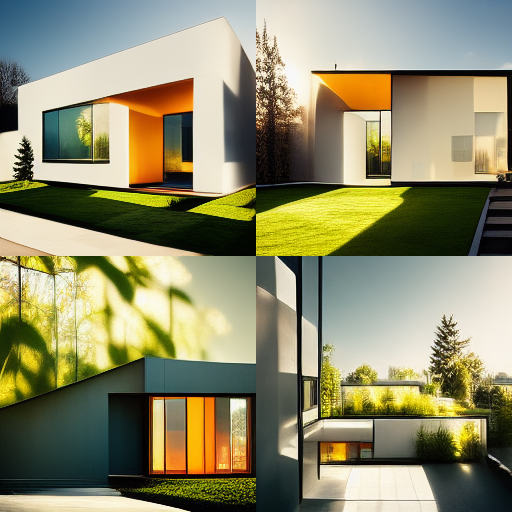} }}%
    \subfloat[\centering Variants]{{\includegraphics[width=0.33\textwidth]{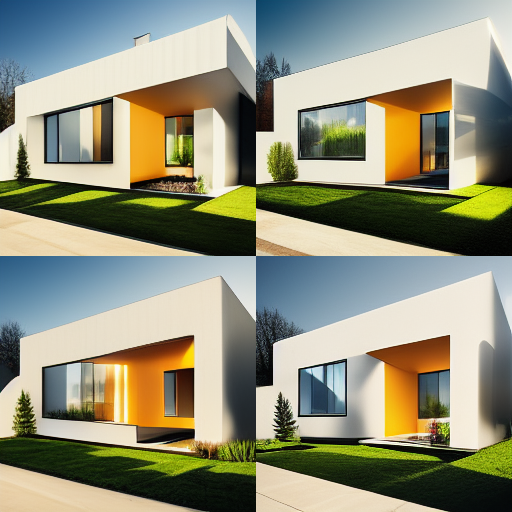} }}%
    \subfloat[\centering Remastered and Upscaled]{{\includegraphics[width=0.33\textwidth]{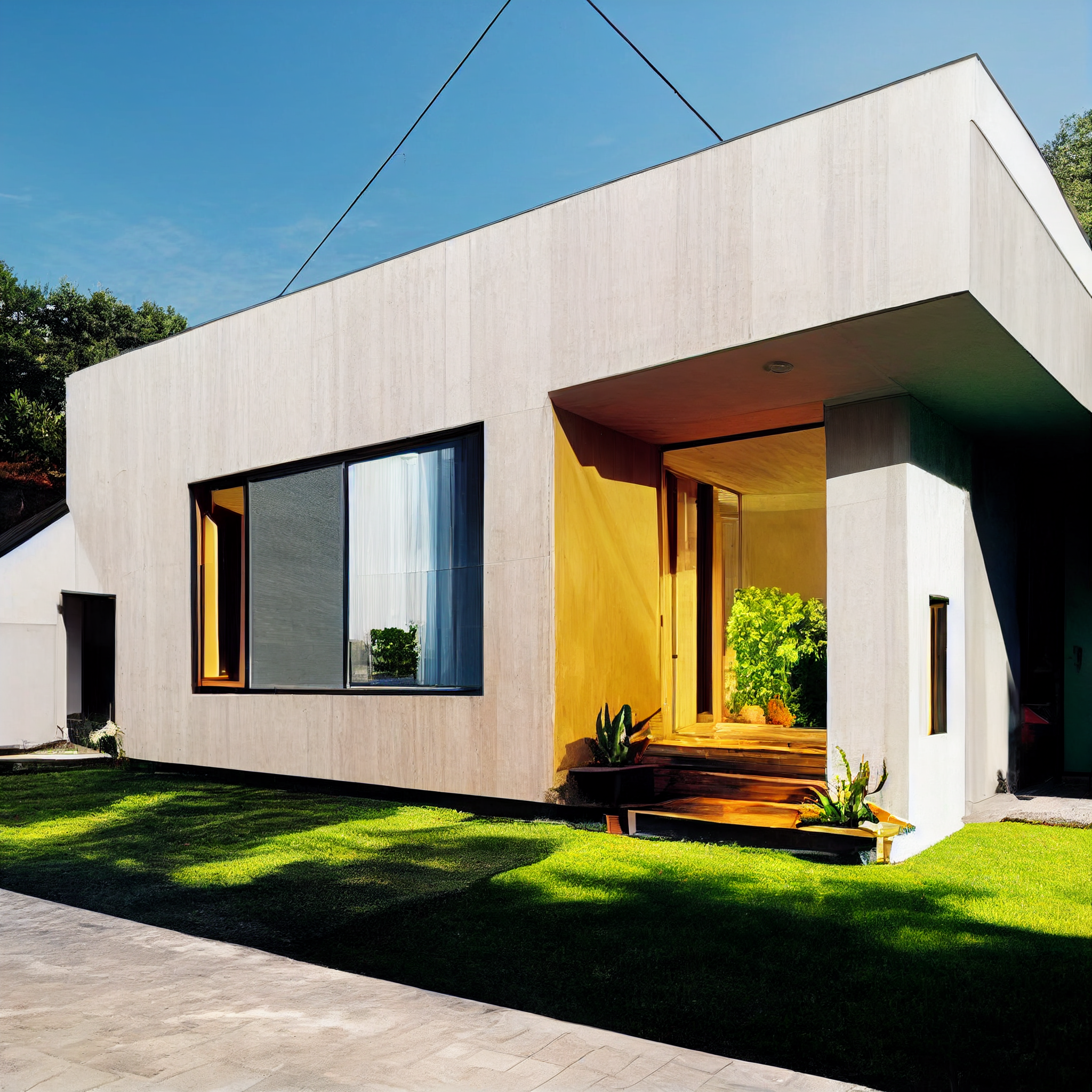} }}%
\vspace{-0.1cm}
    
    \subfloat[\centering DALL-E outpainting of (c)]{{\includegraphics[width=0.33\textwidth]{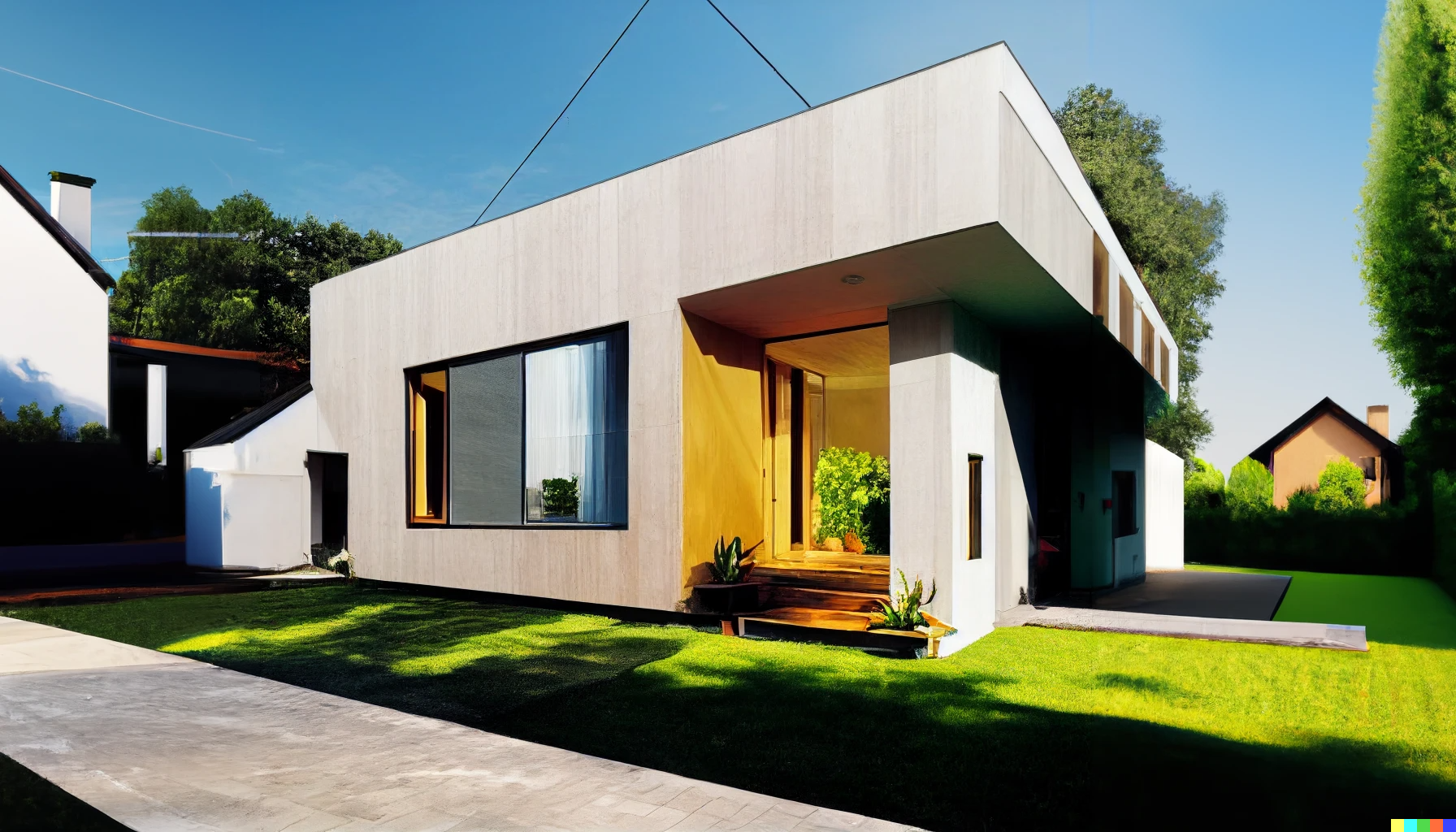} }}%
    \subfloat[\centering Erased parts]{{\includegraphics[width=0.33\textwidth]{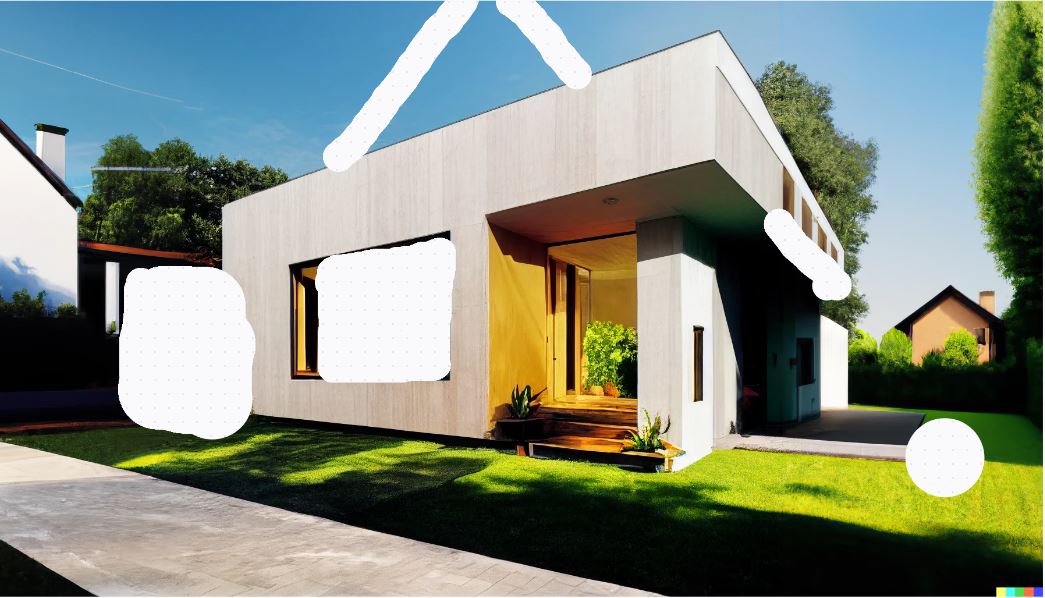} }}%
    \subfloat[\centering Inpainted]{{\includegraphics[width=0.33\textwidth]{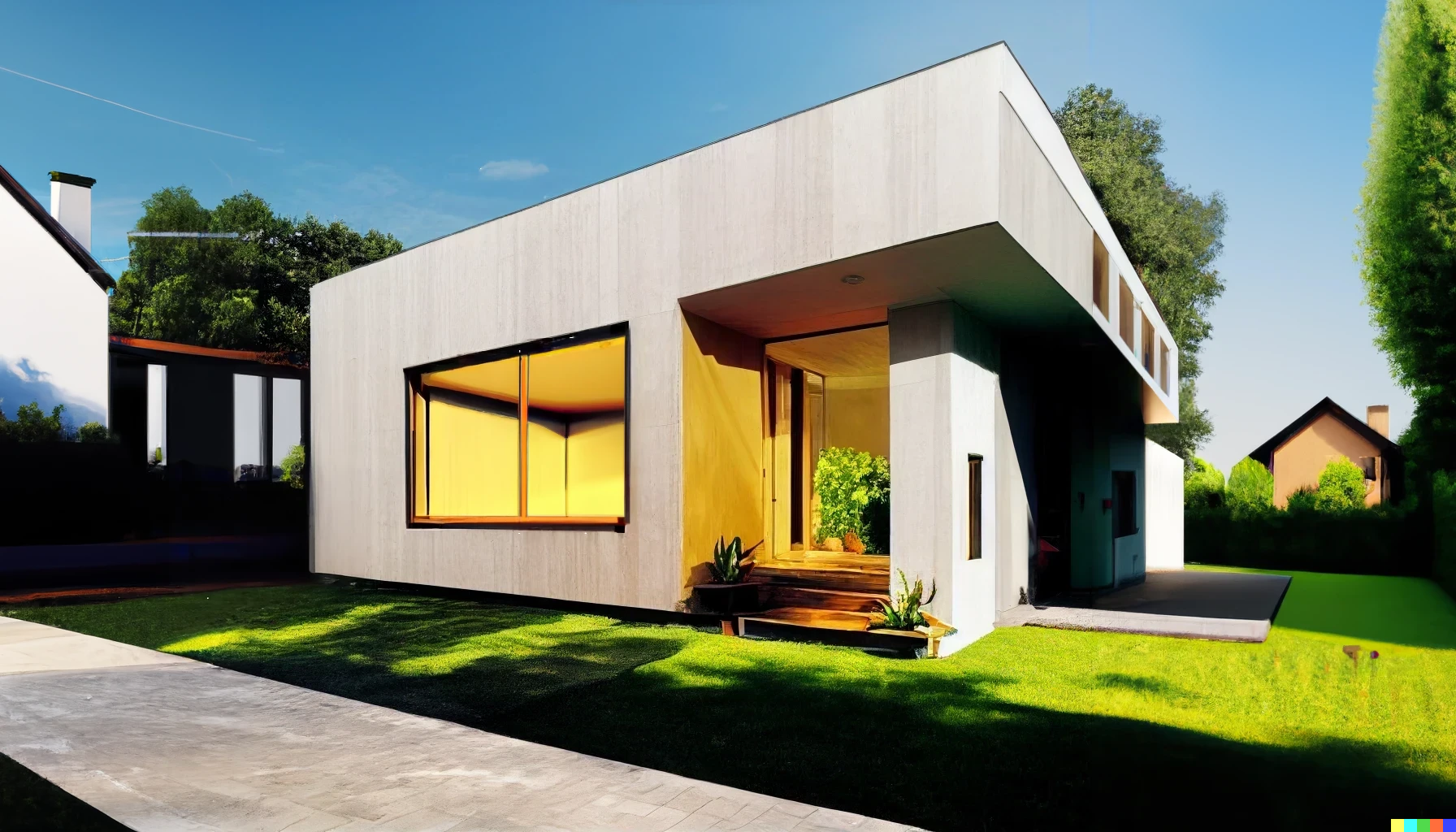} }}%
\vspace{-0.1cm}
    
    \subfloat[\centering Stable Diffusion inpainted walkway]{{\includegraphics[width=0.33\textwidth]{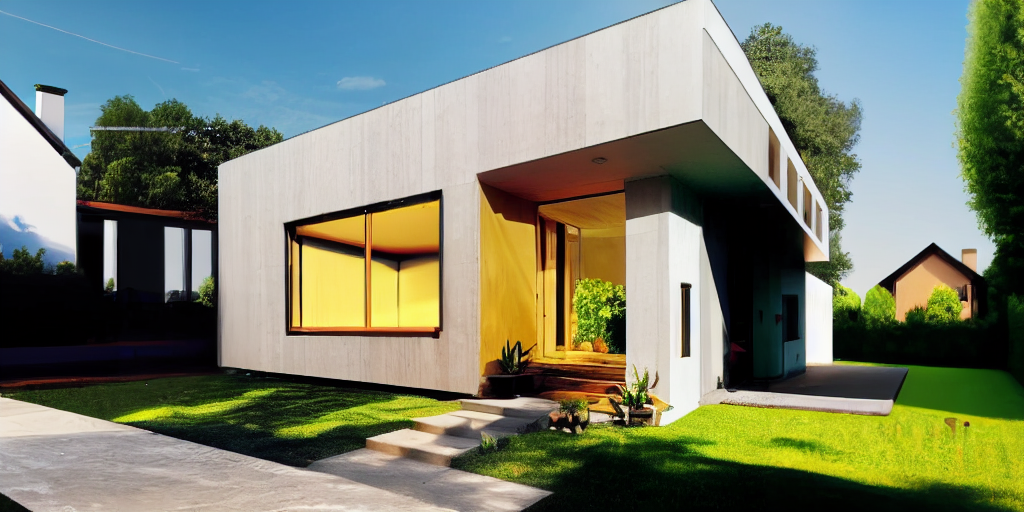} }}%
    \subfloat[\centering 2$^{nd}$ story variant 1]{{\includegraphics[width=0.33\textwidth]{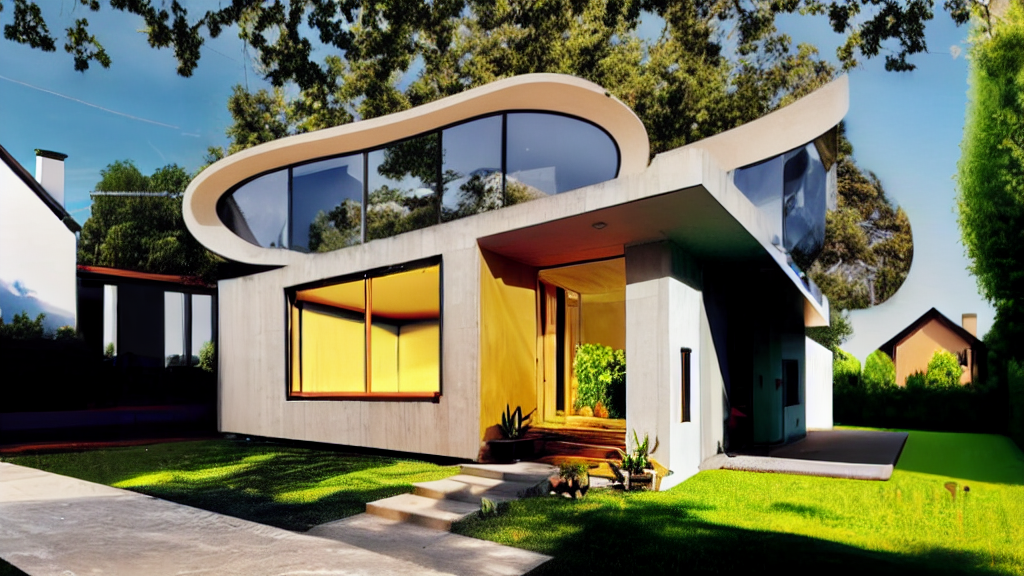} }}%
    \subfloat[\centering 2$^{nd}$ story variant 2]{{\includegraphics[width=0.33\textwidth]{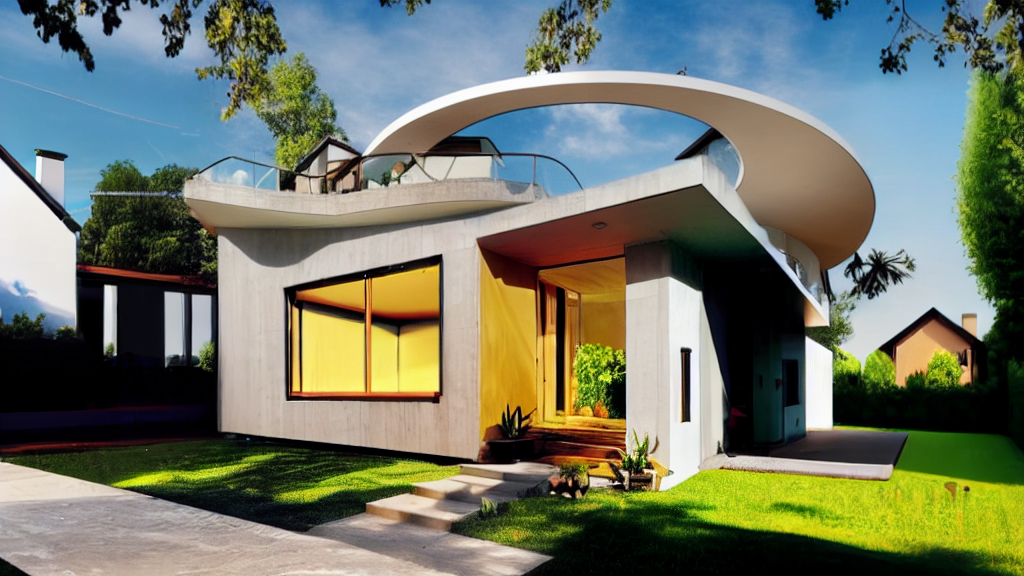} }}%
\vspace{-0.1cm}

    \caption{Refinement and variant generation in Midjourney (a)-(c), DALL-E 2 (d)-(f), and StableDiffusion for a walkway (g) and a second story (h)-(i)}%
    \label{fig:exterior}%
\end{figure}

For DALL-E and StableDiffusion, iteration over image variants is usually a shorter, more precise process as iterations require direct intervention on the image data in form of in-/outpainting. Once we understand these tools and strengths of the platforms, we can use them to combine them into a more flexible workflow. In the following ideation workflow, we will start with a Midjourney prompt to create a desired scene (as it excels at free-form ideations), which we will then refine through in-/outpainting in DALL-E and Stable Diffusion to attain a specific result inspired by the original Midjourney result. An overview of the workflow is shown in Fig.~\ref{fig:combination_workflow} that we will explain along the results in Fig.~\ref{fig:exterior}.

Fig.~\ref{fig:exterior} (a)-(c) shows the beginning stages of an idea as generated in Midjourney. The query that led to this particular result was \enquote{single-family home with garden, full exterior view, modern architecture, photo, sunlight}. Multiple keyword arrangements and weights on different terms were tried before this result was selected, remastered and then upscaled. It is also possible to start with one or more reference images, though that technique was omitted here.

The result was then uploaded to the DALL-E, where it was outpainted in Fig.~\ref{fig:exterior} (d) to create a wider viewing angle and subsequently inpainted in Fig.~\ref{fig:exterior} (e)-(f) to replace unwanted details like the cables hanging in the air or the differently colored windows.

After unsuccessful attempts to add a paved path from the sideway to the entrance through inpainting in DALL-E, we transferred image (f) to StableDiffusion. We erased the walkway and replaced it with inpainting using a modified query without a garden reference (as the walkway would often be overgrown by plants) and adding "paved" early in the prompt (as earlier keywords are weighted higher). This resulted in the prompt \enquote{single-family home, paved between sidewalk and door, full exterior view, modern architecture, photo, sunlight} with the result in image (g).

If we now imagine a situation where the product of this process is used to present to a possible client, requests for variations need to be accommodated. Lets imagine the case that a client asks for a second story or some other roof element. This can easily be accomplished via inpainting by erasing the roof and part of the sky and a slightly changed prompt specifying the style of the new image. Fig.~\ref{fig:exterior} (h) and (i) show the result for the StableDiffusion query \enquote{single-family home, two stories, clear blue sky, curved roof, full exterior view, modern architecture, photo, sunlight} after the roof and central area of the sky have been erased and the canvas has been extended upwards to give more room for roof elements. Note that some latent effects like the tree branches reaching into the image are hard to avoid.

\subsection{Common Limitations}

Trial and error is a reality of working with current AI models. They usually do not output the desired results right on the first try. Instead they offer multiple variants for each query and expect the user to pick the best one or refine their wordings until they arrive at an acceptable result. This may not be straight forward, but giving a group of architecture students the task to draw a design with similarly rough specifications would also result in many variants and would be way more time consuming.

There are however cases where the variant tree that users explore simply fails entirely and a good result will never be reached without excessive amounts of trial and error, retraining or inpainting.

\begin{figure}[t]%
    \centering

    \subfloat[\centering \enquote{architectural floor plan of a modern single-family home, ground floor}]{{\includegraphics[width=0.33\textwidth]{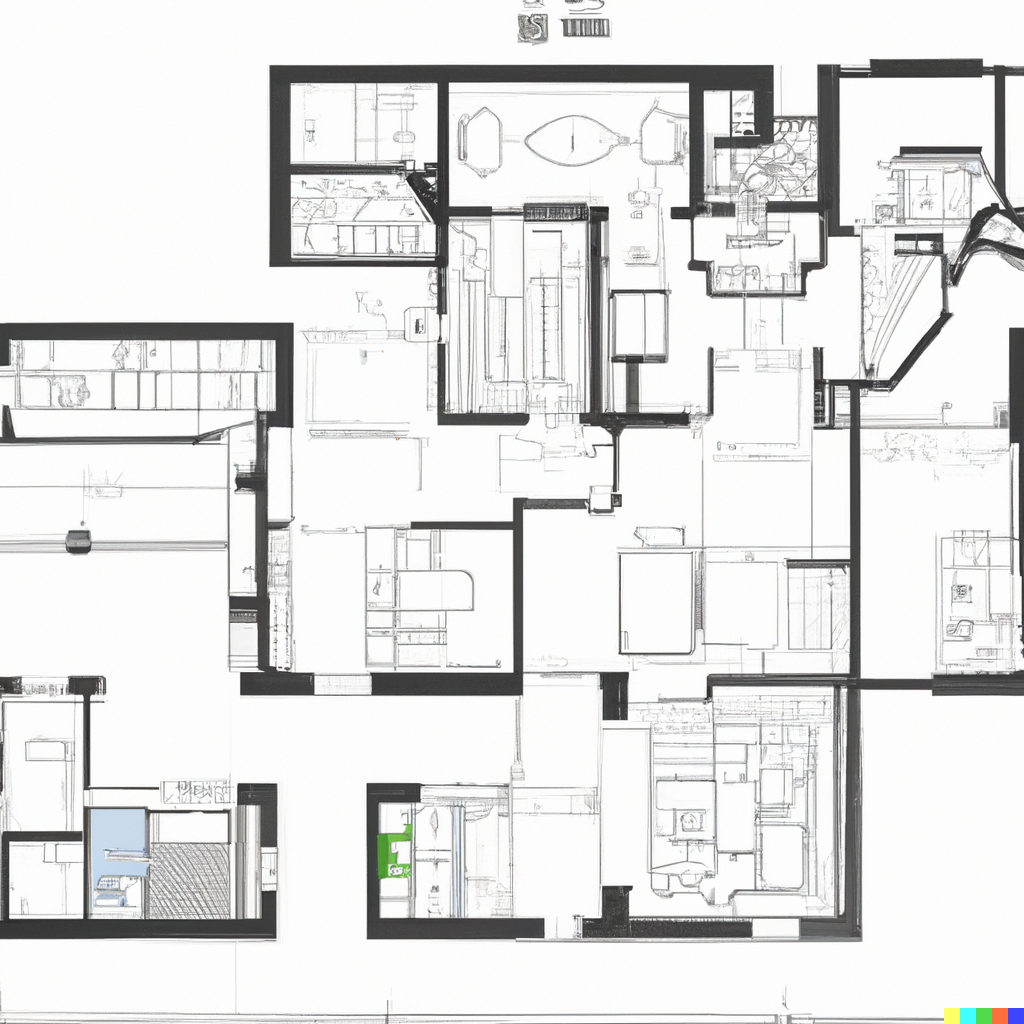} }}%
    \subfloat[\centering \enquote{red brick building facade with a dutch clock gable and a bow window}]{{\includegraphics[width=0.33\textwidth]{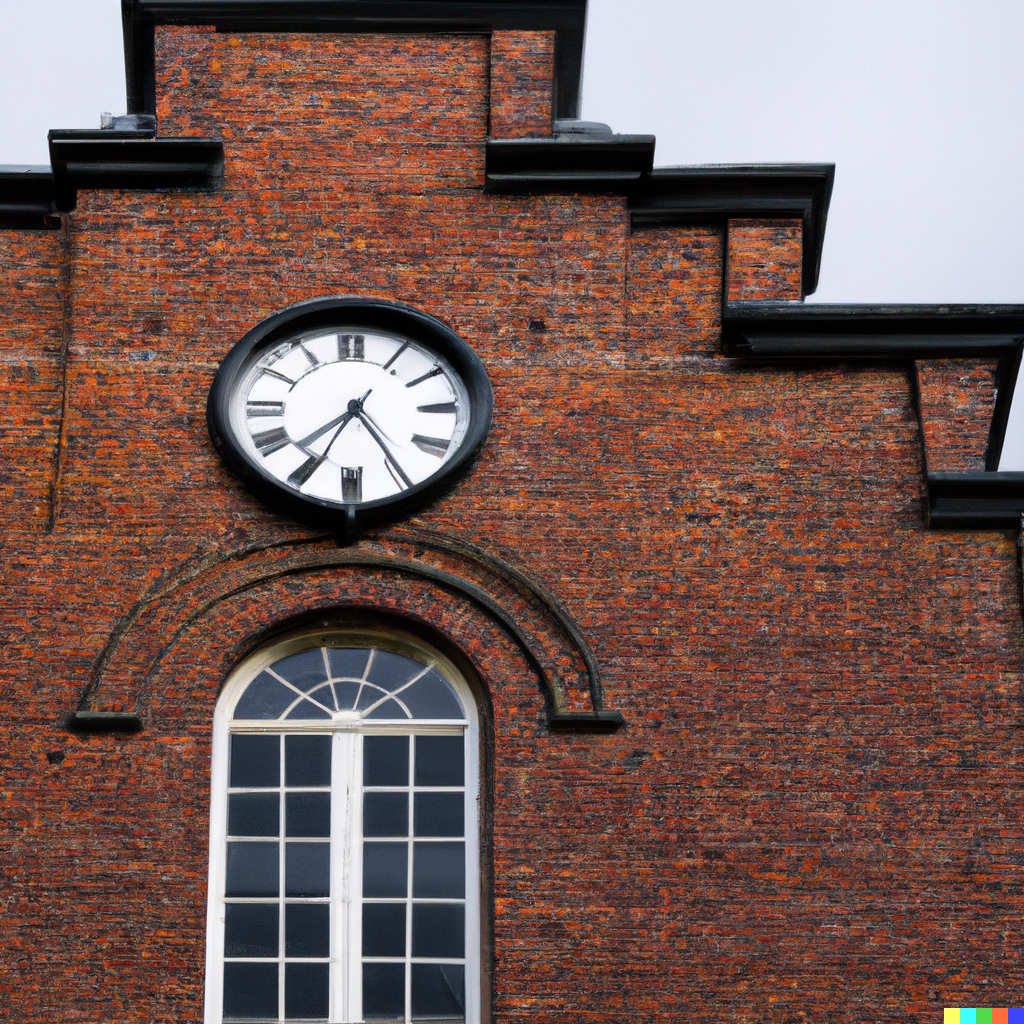} }}%
    \subfloat[\centering \enquote{a drawing of a map of five buildings arranged around a park, realistic, detailed, top-down, site plan, digital rendering}]{{\includegraphics[width=0.33\textwidth]{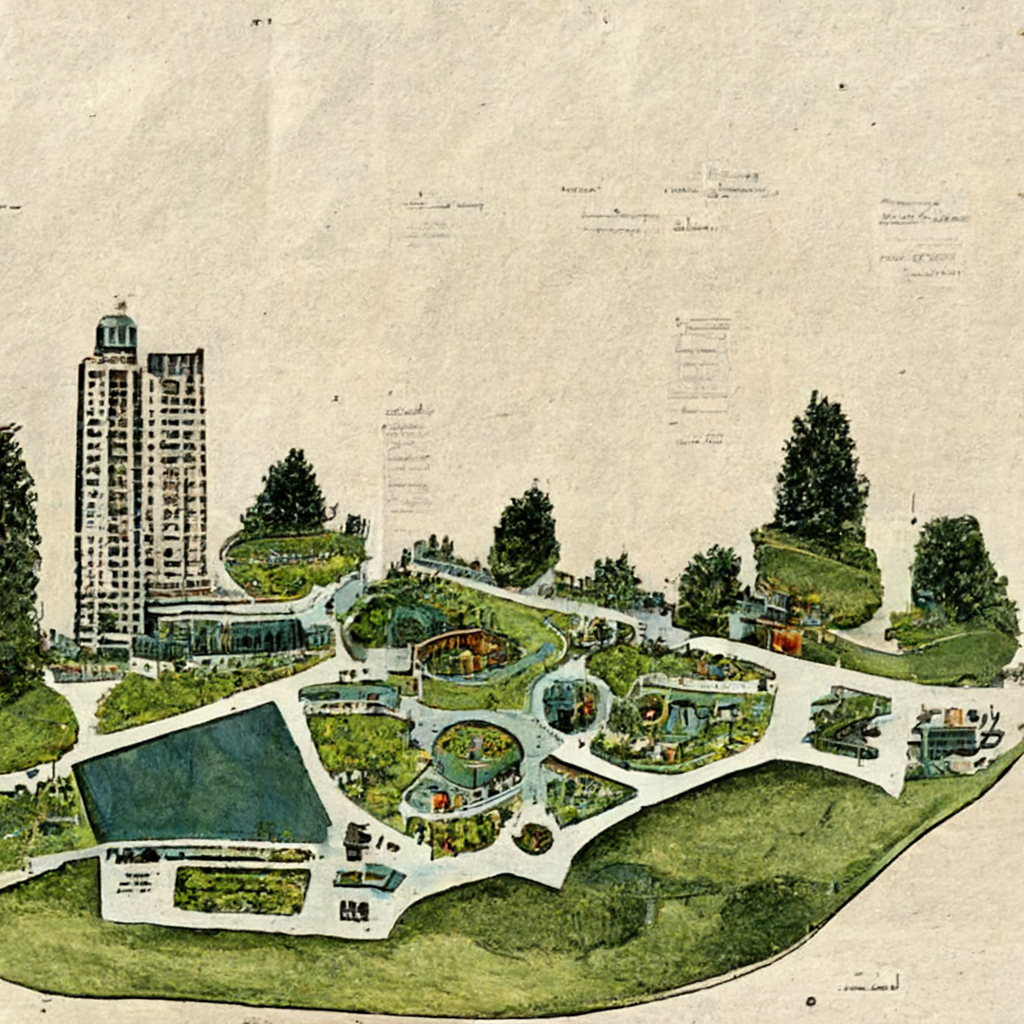} }}%
    
    \caption{Example failure cases.}%
    \label{fig:failure}%
\end{figure}

Some of the common failure cases an architect would encounter while using these models are shown in Fig.~\ref{fig:failure}. Case (a) shows the result for a floor plan query. It does well in imitating the style of bold lines for walls and thin lines for objects, but becomes completely nonsensical. This is due to the fact that AI art tools replicate the style, but have no semantic understanding of the meaning of the lines in a floor plan. Case (b) shows the result of a query with multiple specific technical terms, which are also somewhat ambiguous in themselves. The specific architectural element \enquote{bow window} was turned into a window with a bow over it. The \enquote{clock gable} was represented by a different gable element with a clock below it. Case (c) shows a query for a landscape design with five buildings, which invokes the common problem that these AI tools are just bad at counting and complex spatial arrangements beyond foreground and background.

\section{Conclusion}\label{sec:conclusion}

In this paper we looked at AI art generation tools through multiple lenses to present a snapshot of the current situation and discuss their applicability to practise. To start out we identified architectural design use cases that can already be tackled with image generation models and use cases that are likely to be unlocked soon. We highlighted the differences between three of the current main contenders. We took a quantitative look by analysing millions of queries at how users are already querying these models for architectural results and how many queries it seemed to take them until a satisfactory result was reached. Finally, we tried to package these insights into workflow examples that readers might adapt in their own experiments with these different platforms.

In the coming months and years these models and platforms will evolve  in their capabilities and new contenders will appear. A trend that we already see is that the workflows across tools move closer together as appropriate interaction paradigms establish. In the end, we will likely have large ready-made platforms that can deliver any style and interaction the user desires and they will work even better for use cases like ideation, collages, build and style variants.

The use cases that are currently beyond these models and platforms all require a semantic understanding of the content of the images. A floor plan is not just a collection of lines. These lines represent semantic and contextual information like that a room actually requires a door to get to another connected room separated by a wall. But, as we already have Building Information Models (BIM) that provide this semantic information it is just a matter of time that new models will arrive that are trained specifically on these datasets and it is likely that they will be able to fulfil all requirements from Table~\ref{tab:usecases_comparative}.

The many use cases and results shown in this paper illustrate a strong potential for AI models in architecture and the \filtersize queries with architectural context that we identified demonstrate already a strong adoption. One of these users, Hassan Ragab, summarizes this potential well on his instagram page\footnote{\url{https://www.instagram.com/p/Cj-zf5IJ6Mt/}}: \enquote{As creatives we always have the moment where we stop. [...] And there lies the true power when working with AI. It allows you to make that push, to add something to your limits in the same way a plane allowed us to fly. It's another barrier that we are breaking.}

\bibliographystyle{elsarticle-num} 
\bibliography{sn-bibliography}

\end{document}